\documentclass[10pt,twocolumn,letterpaper]{article}

\usepackage{cvpr}              
\usepackage{float}
\usepackage[dvipsnames]{xcolor}
\usepackage{multirow}
\usepackage{bbm}

\newcommand{\figref}[1]{Fig.~\ref{#1}}
\newcommand{\secref}[1]{Section~\ref{#1}}

\newcommand{\tabref}[1]{Tab.~\ref{#1}}

\newcommand{\suppl}{\textcolor{black}{SupMat}\xspace}
\definecolor{Green4}{RGB}{1,50,32}
\newif\ifshowcomments
\showcommentstrue 

\ifshowcomments
        \newcommand{\note}[3]{{\textcolor{#2}{[#1: #3]}}}
        \newcommand{\OH}[1]{\note{Otmar}{green}{#1}}
	\newcommand{\SC}[1]{\note{Sammy}{teal}{#1}}
        \newcommand{\ZF}[1]{\note{Alex}{blue}{#1}}
	\newcommand{\JS}[1]{\note{Jie}{magenta}{#1}}
        \newcommand{\HZ}[1]{\note{Hui}{orange}{#1}}
\else
        \newcommand{\note}[3]{\unskip}
        \newcommand{\OH}[1]{\unskip}
	\newcommand{\SC}[1]{\unskip}
	\newcommand{\JS}[1]{\unskip}
	\newcommand{\ZF}[1]{\unskip}
	\newcommand{\HZ}[1]{\unskip}
\fi

\newcommand{\dgrasp}{\mbox{D-Grasp}\xspace}

\newcommand{\manipnet}{\mbox{ManipNet \cite{zhang2021manipnet}}\xspace}
\newcommand{\imos}{\mbox{IMoS \cite{ghosh2022imos}}\xspace}

\newcommand{\dexmv}{\mbox{DexMV \cite{qin2022dexmv}}\xspace}

\newcommand{\arctic}{\mbox{ARCTIC \cite{fan2023arctic}}\xspace}
\newcommand{\cams}{\mbox{CAMS \cite{zheng2023cams}}\xspace}

\newcommand{\methodname}{ArtiGrasp}

\newcommand{\TITLE}{\methodname: Physically Plausible Synthesis of Bi-Manual Dexterous \\ Grasping and Articulation}

\newcommand{\fullbody}{full-body\ }
\newcommand{\task}{\textit{Dynamic Object Grasping and Articulation}}

\newcommand{\greencheck}{{\color{Green4} \checkmark}}
\newcommand{\redcross}{{\color{red}$\times$}}

\newcommand{\colorRef}[1]{\textcolor{black}{#1}} 

\newcommand{\refTab}[1]{\mbox{\colorRef{Table~\ref{#1}}}}

\newcommand{\heading}[1]{\textbf{#1}.}

\newcommand{\sixD}{{6D}\xspace}

\newcommand{\threeD}{\xspace{3D}\xspace}
\newcommand{\actions}{\mathbf{a}}
\newcommand{\states}{\mathbf{s}}
\newcommand{\torques}{\boldsymbol{\tau}}
\newcommand{\policyvec}{\boldsymbol{\pi}}

\definecolor{Gray}{gray}{0.9}

\newcommand{\V}[1]{\mathbf{#1}} 
\newcommand{\R}{\rm I\!R}

\renewcommand{\thefootnote}{\fnsymbol{footnote}}

\makeatletter
\def\and{
  \end{tabular}%
  \hskip 2.85em \@plus.17fil%
  \begin{tabular}[t]{c}}
\makeatother

\definecolor{cvprblue}{rgb}{0.21,0.49,0.74}
\usepackage[pagebackref,breaklinks,colorlinks,citecolor=cvprblue]{hyperref}

\def\paperID{253} 
\def\confName{3DV\xspace}
\def\confYear{2024\xspace}

\title{\TITLE}

\author{%
  Hui Zhang$^{1,2*}$ \and Sammy Christen$^{1*}$ \and Zicong Fan$^{1,2}$ \and Luocheng Zheng$^1$ \and Jemin Hwangbo$^{3}$ \and Jie Song$^1$ \and Otmar Hilliges$^1$ 
  \vspace{1mm}
  \and
    $^1$ETH Zurich \quad $^2$Max Planck Institute for Intelligent Systems \\ $^3$Korea Advanced Institute of Science and Technology (KAIST) \vspace{-1mm} \\ \vspace{-3mm}
  }

\begin{document}

\twocolumn[{%
\maketitle
\begin{figure}[H]
\hsize=\textwidth 
\centering
\vspace{-10mm}
\includegraphics[width=0.95\textwidth]{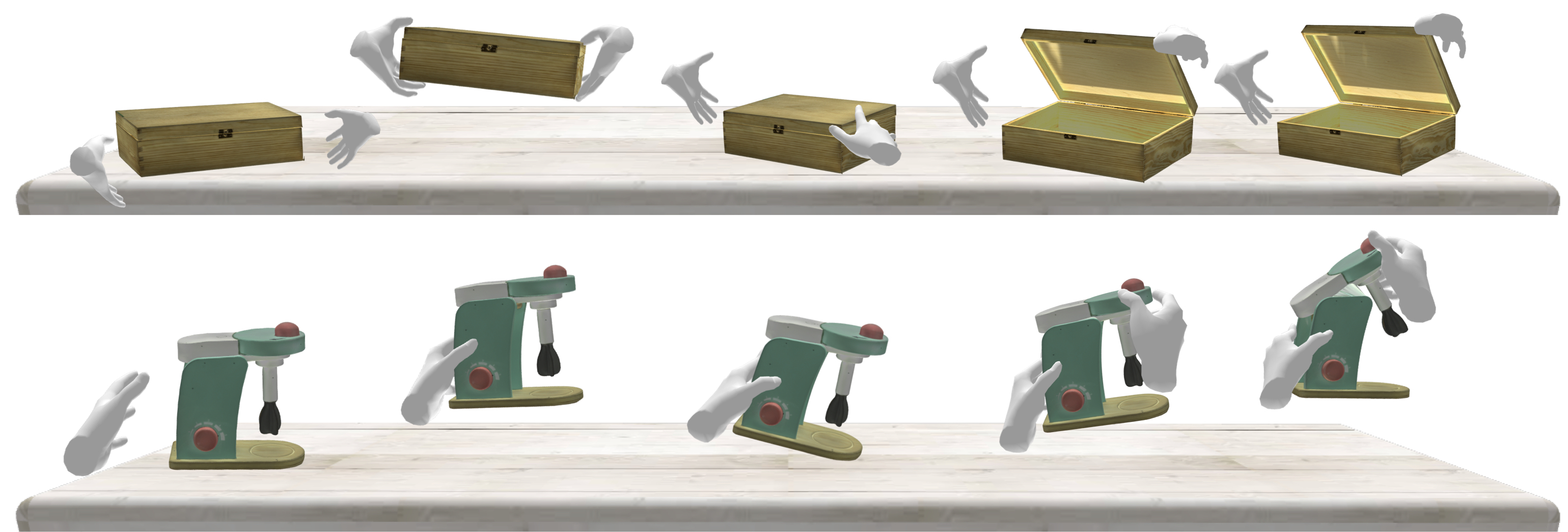} 
\caption{We present a method to synthesize physically plausible bi-manual manipulation. Our method can generate motion sequences such as grasping and relocating an object with one or two hands, and opening it to a target articulation angle.}
\label{fig:teaser}
\end{figure}
}]

\maketitle
\def\thefootnote{*}\footnotetext{These authors contributed equally to this work 

Correspondence: huizhang@ethz.ch}

\begin{abstract}
\vspace{-2mm}
We present \methodname, a novel method to synthesize bi-manual hand-object interactions that include grasping and articulation. This task is challenging due to the diversity of the global wrist motions and the precise finger control that are necessary to articulate objects. \methodname\ leverages reinforcement learning and physics simulations to train a policy that controls the global and local hand pose. Our framework unifies grasping and articulation within a single policy guided by a single hand pose reference. Moreover, to facilitate the training of the precise finger control required for articulation, we present a learning curriculum with increasing difficulty. It starts with single-hand manipulation of stationary objects and continues with multi-agent training including both hands and non-stationary objects. To evaluate our method, we introduce \task, a task that involves bringing an object into a target articulated pose. This task requires grasping, relocation, and articulation. We show our method's efficacy towards this task. We further demonstrate that our method can generate motions with noisy hand-object pose estimates from an off-the-shelf image-based regressor.  Project page: \href{https://eth-ait.github.io/artigrasp/}{https://eth-ait.github.io/artigrasp/}.
\end{abstract}
    
\section{Introduction}
\label{introduction}

The ability to manipulate complex objects, such as operating a coffee machine, opening a laptop, or passing a box, is a fundamental part of everyday life. Providing systems with the capability to understand and perform such tasks can enable effective interactions with the physical world and provide assistance to humans in various domains. Consequently, the capacity to generate realistic hand-object interactions is paramount in fields like animation, AR/VR, human-computer interaction, and robotics. Traditional methods for capturing human motion in gaming and films, such as multi-view marker-based setups, can be costly and require substantial data cleaning for motion capture data~\cite{fan2023arctic,taheri2020grabnet}. Hence, a model that can proficiently generate two-handed motions interacting with objects could reduce the costs associated with motion capture.

Research has turned to synthesizing hand-object interactions using either data-driven~\cite{zhang2021manipnet,zheng2023cams} or physics-based methods~\cite{christen2022dgrasp}. While existing data-driven methods generate hand-object motions, including object articulation~\cite{zheng2023cams} and two-hand manipulation~\cite{zhang2021manipnet}, these methods typically depend on complete supervision from precise 3D motion data for each frame (see \refTab{tab:related_work}). Recently, physics-based methods that leverage reinforcement learning (RL) in a simulated environment have been proposed~\cite{christen2022dgrasp}. This approach reduces the data requirement for motion generation as they demand only a single hand pose reference per interaction. While physics-based approaches have primarily focused on single-hand grasping motions for rigid objects, real world hand-object interactions are often bi-manual and include articulation. However, a framework for synthesizing bi-manual grasping and articulation of objects is still missing. 

Here, we go beyond single-hand grasping interaction of rigid objects and present~\methodname, a novel method to synthesize dynamic bi-manual grasping and articulation of objects. We formulate this task as a reinforcement learning problem and leverage physics simulations. This allows our method to learn motions that adhere to physical plausibility, ensuring no object interpenetration and that object articulation results from stable hand-object contacts and forces. We propose a general reward function and training scheme that enables grasping and articulation of a diverse set of objects without object- or task-specific retraining. 

Object articulation and bi-manual grasping present two key challenges compared to single-hand grasping. First, the articulation of different objects requires diverse wrist motions, making it challenging to define a general control strategy. For example, we show that a simple PD control scheme for relocation of objects after grasping~\cite{christen2022dgrasp} does not work well in this setting. To address this, we train an RL-based policy that learns to i) manipulate an object to a target articulation angle and ii) achieve natural interactions with the objects by utilizing only a single hand pose reference as input. The second key challenge is the precise finger control that is necessary to achieve successful articulation, where even small deviations from ideal positions on the target object impact performance. In the bi-manual manipulation setting, one hand can easily hinder the other hand from reaching its ideal position. To deal with this challenge, we introduce a learning curriculum consisting of two phases. In the first phase, we fix the object base to the surface and create separate learning environments for each hand. This allows our policies to focus on learning precise finger control for articulation. In the second phase, we fine-tune the policies using non-fixed objects in a shared physics environment, allowing the hands to cooperate. 

In our experiments, we first assess both grasping and articulation separately, and then evaluate the \task~task, which involves transitioning an articulated object from its initial state into a target articulated object pose (see \figref{fig:teaser}). To the best of our knowledge, there are no direct baselines for this task. Therefore, we adjust the closest related work~\cite{christen2022dgrasp} and show that simple adaptations lead to low task success rates. On the other hand, our method achieves performance gains of 5$\times$ over this baseline. We further demonstrate that our method can work with inputs from both motion capture data and noisy reconstructed poses from images with off-the-shelf hand-object pose estimation models. Lastly, we ablate the main components of our framework.

Our contributions can be summarized as follows: 1) We propose a method to achieve \task~in a physically plausible manner. 2) Our method leverages RL and a general reward function to learn fine-grained wrist and finger control to grasp and articulate different objects without task- or object- specific retraining. 3) We present a learning curriculum with increasing difficulty to address the complexity of learning articulation and bi-manual manipulation. 4) We demonstrate that our method can utilize hand pose estimates from a single image as input to generate dynamic grasping and articulation. 
\section{Related Work}
\label{related}

\begin{table}
\centering
\begin{center}
\resizebox{0.45\textwidth}{!}{%
\begin{tabular}{l|cccc}
   \toprule
        & \multirow{2}{*}{Few Shot}  & Physics  & \multirow{2}{*}{Two-hand}  & Free-base \\
     Method      &   & Simulation  & & Articulation  \\
    \midrule
    \imos    & \redcross    & \redcross                 & \greencheck & \redcross    \\
    \manipnet  & \redcross   & \redcross                 & \greencheck & \redcross \\
    \cams       & \redcross  & \redcross                & \redcross & \greencheck \\
    Zhang \etal \cite{zhang2023learning} & \redcross & \greencheck             & \redcross   & \redcross   \\
    \dexmv       & \redcross & \greencheck               & \redcross   & \redcross   \\
    \dgrasp \cite{christen2022dgrasp}     & \greencheck & \greencheck           & \redcross   & \redcross   \\
    \textbf{\methodname\ (Ours)} &  \greencheck & \greencheck         & \greencheck & \greencheck  \\
   \bottomrule
\end{tabular}
}
\end{center}
\vspace{-0.25cm}
\caption{\heading{Comparison between ours and existing methods} Ours generates two-hand manipulations using physics simulation, requires only static hand pose references (few shot), and accommodates both rigid and articulated objects with a unified policy.}
\label{tab:related_work}
\vspace{-0.2cm}
\end{table}

We categorize related research into human grasp generation, motion synthesis, and dexterous robotic hand manipulation. \refTab{tab:related_work} compares different hand-object motion generation methods to ours. 

\subsection{Hand-object Reconstruction and Synthesis}

There has been a surge of large-scale datasets introduced for modeling hand-object interaction~\cite{brahmbhatt2019contactdb, hampali2020ho3d, brahmbhatt2020contactpose, chao2021dexycb, kwon2021h2o, wang2023dexgraspnet}. Recent datasets include interactions with articulated objects~\cite{fan2023arctic,zhu2023contactart,liu2022ho4d,xu2021d3d}. Methods that are being developed with these datasets can be split into two categories: 1) \textit{hand-object reconstruction}; 2) \textit{hand grasp synthesis}.

The goal of \textit{hand-object reconstruction}~\cite{hasson2019obman,liu2021semi,Yang_2021_CPF,grady2021contactopt,Hasson2020photometric,tekin2019ho,corona2020ganhand,zhou2020monocular, zhou2022toch, ziani2022tempclr} is to estimate the \threeD hand and object surfaces from RGB images. Methods mostly leverage deep learning models, for example, by using 3D supervision and synthetic images~\cite{hasson2019obman}, temporal models~\cite{tekin2019ho}, or by integrating adversarial priors and contact constraints~\cite{corona2020ganhand}. Others focus on the denoising of pose estimates~\cite{grady2021contactopt, zhou2022toch}.

In \textit{hand grasp synthesis}, the goal is to generate static hand grasps given an object mesh as input. 
Corona \etal~\cite{corona2020ganhand} first generate the grasp type and then refine the grasp pose accordingly. 
Karunratanakul \etal~\cite{karunratanakul2021halo} introduce a part-based implicit hand model for grasp synthesis.
Hidalgo-Carvajal \etal~\cite{hidalgo2022objectcentric} use 10 pre-defined poses to predict infeasible grasping areas and feasible hand poses.
Some methods use contact information to predict or refine the grasp configuration~\cite{jiang2021graspTTA, li2023contact2grasp, ye2012synthesis, yang2021ArtiBoost, grady2021contactopt}. 
Moreover, physics-based optimization can be leveraged to improve the initial grasping pose~\cite{wang2023dexgraspnet, wu2022learning}. 
Han \etal~\cite{han2023vrhand} optimize a single reference grasp pose for grasping different objects in virtual reality. 
Some other approaches focus on more general tasks, such as generating grasps for dexterous hands and grippers~\cite{li2023gendexgrasp} or using implicit representations~\cite{karunratanakul2020grasping} for grasping and reconstruction.

The methods above are orthogonal to ours.
They do not generate hand motion but provide useful static grasp poses that we can potentially leverage to synthesize motions involving grasping and manipulating objects.

\subsection {Motion Synthesis}
The synthesis of human motion is a long-standing problem in graphics and computer vision~\cite{ Aksan_2019_ICCV, aksan2020motiontransformer, ghosh2017learning,holden2015tog, holden2016tog}. 
Recently, there have been methods that synthesize human body motion interacting with static scenes~\cite{hassan2021samp, cao2020long, starke2020local,zhang2022couch,wang2021scene,lee2023locomotionactionmanipulation, huang2023diffusionbased, zhang2023roam}, such as sitting on a chair or sofa.
This line of work focuses on the human body motion and does not generate motions of hand-object interaction.
There are methods that generate \fullbody grasping motion~\cite{taheri2021goal, ghosh2022imos, li2023taskoriented, wu2022saga}, but they are purely data-driven approaches without physics. They either require a post-processing optimization for the object motion~\cite{ghosh2022imos}, or only generate the approaching motion until grasping~\cite{wu2022saga, taheri2021goal}. Recent work has also explored integrating physics simulations into pose reconstruction~\cite{yuan2021simpoe, shimada2020physcap, shimada2021neural}, synthesis~\cite{xie2021iccv}, and even coarse human-object interaction pipelines~\cite{chao2021learning,luo2022embodied,hassan2023synthesizing}. However, these physics-based methods focus on the body motion and do not model hands to capture fine-grained hand-object interactions. In contrast, we omit the human body and focus specifically on bi-manual hand-object manipulation.

Similar to ours, some recent methods also focus on dexterous hands~\cite{wang2019learning} and generate interactive sequences that include physics~\cite{christen2022dgrasp, braun2023physically}, articulated objects~\cite{zheng2023cams}, or bi-manual manipulation~\cite{zhang2021manipnet}. Wang \etal~\cite{wang2019learning} generate multi-step human-object interactions from videos without physics. Christen \etal~\cite{christen2022dgrasp} generate natural grasping sequences from static grasp references in a physics simulation, but focus on rigid objects and only consider one hand. Zheng \etal~\cite{zheng2023cams} presents a data-driven method to synthesize single-hand pose sequences for articulated objects and uses post-optimization to make the sequences more physically plausible. 
Zhang \etal~\cite{zhang2021manipnet} predict finger poses for two-hand object interaction. However, they require the full trajectory of the wrist and object at inference time. 
In our work, we generate object and wrist motions through interaction with the objects in a physics simulation.
 
\subsection{Dexterous Robotic Hand Manipulation}
Dexterous robotic manipulation is often addressed by leveraging physics simulation and reinforcement learning with robotic hands~\cite{she2022grasping, ye2023grasping, liu2023dexrepnet, rajeswaran2018learning, garciahernando2020iros}. Some common strategies to learn to grasp different objects include data augmentation~\cite{Chen2022LearningRR}, curriculum learning~\cite{zhang2023learning}, or improving the movement re-targeting from human demonstrations to the robot hand in a physics simulation~\cite{qin2022dexmv, chen2022dextransfer, liu2023dexrepnet}. Other approaches include tele-operated sequences as training data~\cite{rajeswaran2018learning} or use demonstrations as prior and predict residual actions~\cite{garciahernando2020iros}. Alternatively, a parameterized reward function based on demonstrations can be used~\cite{christen2019hri}. All of these methods rely on demonstration of entire grasping sequence. In contrast, our method only requires static hand pose references, which are easier to obtain. 

To alleviate the need for expert demonstrations, some methods are formulated as pure RL problems~\cite{openai2018dexterous, chen2021simple}. Chen \etal~\cite{chen2022humanlevel} propose a benchmark for two-hand manipulation tasks and use standard RL algorithms. However, the hand poses remain rather unnatural. They mitigate this by fine-tuning with a human preference reward~\cite{ding2023learning}. 
Mandikal and Grauman~\cite{mandikal2020graff} train an affordance-aware policy for grasping. They further introduce a hand pose prior learned from YouTube videos to achieve more natural hand configurations~\cite{mandikal2021dexvip}. However, they only consider a single pose per object. Qin \etal~\cite{qin2023dexpoint} achieve grasping objects or opening doors with noisy point clouds from a single camera. Xu \etal~\cite{xu2023unidexgrasp} and Wan \etal~\cite{wan2023unidexgrasp} generate grasp sequences on objects with point clouds. Yang \etal~\cite{yang2022chop} focus on planning interactions with chopsticks that are already in the hand from the start. All these methods consider a single robotic hand and focus mostly on rigid objects, while our method can handle two hands for both grasping and articulation of different objects with a single policy. Chen \etal~\cite{NEURIPS2022BidexHand} propose a benchmark for bi-manual manipulation with different rewards and separately trained policies for each individual task and object. In contrast, our method learns a general policy across different articulated objects for approaching, grasping and articulating.

\begin{figure*}
  \includegraphics[width=0.86\textwidth]{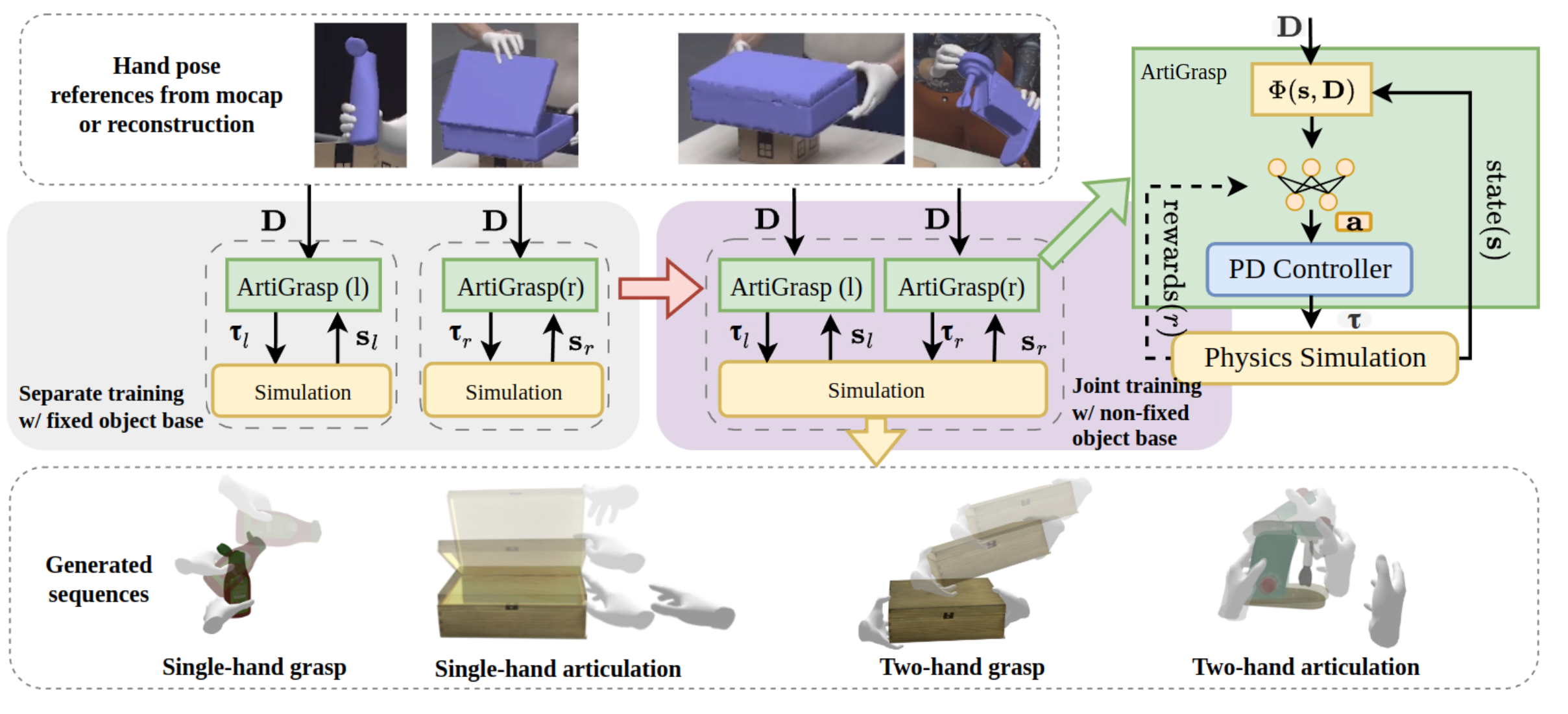}
  \centering
  \vspace{-0.2cm}
  \caption{\heading{Overview of Grasping and Articulation Policy} Our method uses static hand pose references as input (top row) and generates dynamic sequences (bottom row, where higher transparency represents further in time). We propose a curriculum that starts in a simplified setting with separate environments per hand and fixed-base objects (gray solid box on the left) and continues training in a shared environment with non-fixed object base (purple solid box in the middle). Our policies are trained using reinforcement learning and a physics simulation. Rewards are only used during training. The detailed structure of our policy is shown on the right.}
  \label{fig:structure}
\vspace{-0.2cm}
\end{figure*}

\section{Task Definition}
\label{sec:task}
The \task~task is illustrated in \figref{fig:comparison}. We are given an articulated object that consist of two parts rotating about an axis $\V{q}_{\text{ax}}$, an initial articulated object pose $\mathbf{\Omega}^0$, a target articulated object pose $\overline{\mathbf{\Omega}}$, 
and two pairs of object-relative hand pose references $\V{D}$ (one for grasping and one for articulation). Our goal is to generate a sequence of one or two hands interacting with the object such that the initial object pose $\mathbf{\Omega}^0$ approaches the target pose $\overline{\mathbf{\Omega}}$. 
An articulated object pose $ \mathbf{\Omega}$ is defined by the 6 DoF global pose of the object base $\textbf{B}$ and the 1 DoF angle $\omega$ of its articulated joint. We define the output sequence as $\{(\mathbf{q}^t_l, \mathbf{T}^t_l, \mathbf{q}^t_r, \mathbf{T}^t_r, \mathbf{\Omega}^t)\}_{t=1}^{T}$, where $T$ is the number of time steps and $ \mathbf{\Omega}^t$ is the articulated object pose at time step $t$. The hand joint rotations and the global 6D hand pose are defined by $\V{q}_h^t$ and $\V{T}_h^t$ where $h \in \{l, r\}$. The hand pose references $\textbf{D}=(\overline{\textbf{q}}_l, \overline{\textbf{T}}_l, \overline{\textbf{q}}_r, \overline{\textbf{T}}_r)$ can be obtained from motion capture or grasp predictions~\cite{fan2023arctic} (see our experiments in \secref{sec:experiment}).  

\section{Grasping and Articulation Policy}
\label{method}
We provide an overview of our policy learning framework in \figref{fig:structure}. Since we formulate the problem identically for both hands, we will omit the notation ``h'' for simplicity in this section.
 \methodname~is reinforcement learning based, and hence takes as input a state $\states$, provided by a physics simulation, and the hand pose reference $\textbf{D}$. A feature extraction layer $\Phi$ transforms these inputs and passes them to our policy network. We train a policy $\policyvec(\actions| \Phi(\textbf{s},\textbf{D}))$ for each hand. 
The policy predicts actions $\actions$ as PD-control targets, from which torques $\torques$ are computed. The torques are applied to our controllable hand model's joints in the physics simulation and the updated state is again fed to our feature extraction layer. In the physics simulation, we create controllable MANO hand models~\cite{MANO:SIGGRAPHASIA:2017} with mean shape following~\cite{christen2022dgrasp}. The models have 51 DoFs each and are represented by the local hand pose $\V{q} \in \R^{45}$ and global pose $\V{T} \in \R^{6}$. The objects are represented by meshes taken from the ARCTIC dataset~\cite{fan2023arctic}. Each mesh is split into a base and an articulation part with a single connecting joint. We now present details about RL, the feature extraction layer, the reward function, and our learning curriculum. 

\subsection{RL Background}
We formulate our problem as a Markov Decision Process (MDP), where the goal is to train a policy $\policyvec$ to maximize the expected future reward 
 $\mathbb{E}_{\xi \sim \policyvec} \left[ \sum_{t=0}^T\mathcal{\gamma}^t r_t \right]$, where $\mathcal{\gamma} \in [0, 1]$ is a discount factor, $r_t$ is the reward of time step $t$, and $\mathcal{\xi}=[(\states_0, \actions_0), \cdots, (\states_T, \actions_T)]$ a trajectory of state and action sequences generated by the policy interacting with the physics simulation. The probability distribution over all trajectories $\mathbf{\xi}$ is given by $ p_{\theta}(\mathbf{\xi})=p(\states_0)\prod_{t=0}^T p(\states_{t+1}|\states_t, \actions_t) \policyvec(\actions_t| \Phi(\textbf{s}_t,\textbf{D}))$, where $p(s_0)$ is the initial state distribution and $p(\states_{t+1}|\states_t, \actions_t)$ are the transitions determined by the physics simulation. From this probability we compute the expectation of the discounted future rewards. The policy $\policyvec$ is represented by a neural network, whose weights are updated during training. Note that we describe the case where the two hand policies are trained in separate environments here (see \secref{curriculum}). In the case of two hands interacting in the same environment, the update of the simulation state is influenced by the actions of both policies. To simplify notation, we omit the time indication from the equations in the following sections.

\subsection{Feature Extraction}
The state $\states$ at a time step entails the current poses of the hands and object, as well as the contacts and forces per hand joint. We convert this information into features for the policy. Since we train a left-hand and a right-hand policy, the feature space is hand-specific, however, the overall structure is identical and defined as follows:

\begin{equation}
    \Phi(\textbf{s},\textbf{D}) = (\mathcal{H}, \mathcal{O}, \mathcal{G}), 
\end{equation}
\noindent where $\mathcal{H}$, $\mathcal{O}$, and $\mathcal{G}$ are the hand features, object features, and goal features, respectively. 

The hand features $\mathcal{H}$ are defined as $\mathcal{H}=(\textbf{q}, \dot{\textbf{q}}, \textbf{f}, \dot{\tilde{\textbf{T}}})$ where $\mathbf{q}$ and $\dot{\mathbf{q}}$ are the hand joints' local rotations and velocities, $\mathbf{f}$ are the net contact forces of each link of the hand, and $\dot{\tilde{\textbf{T}}}$ are the hand's linear and angular velocities in object-relative frame.

The object features are $\mathcal{O} = (\tilde{\mathbf{\Omega}}, \dot{\tilde{\mathbf{\Omega}}}, \textbf{I}_{\text{art}})$. 
The terms $\tilde{\mathbf{\Omega}}$ and $\dot{\tilde{\mathbf{\Omega}}}$ indicate the articulated object's 7 DoF pose and velocity expressed in wrist-relative frame. We convert global information into wrist-relative features (denoted by $\tilde{\cdot}$) to make the policy independent of the global state and prevent overfitting. 
To provide more information about the object's state with regards to articulation to our policy, we introduce the term $\textbf{I}_{\text{art}}=( \tilde{\textbf{q}}_{\text{ax}}, \tilde{\textbf{q}}_{\text{art}}, l_{\text{art}}, m_{\text{art}}, m_{\text{base}}, )$, where $\tilde{\textbf{q}}_{\text{ax}}$ and $\tilde{\textbf{q}}_{\text{art}}$ are the direction vector of the articulation axis and the direction vector from wrist to the axis, represented in wrist-relative frame. The terms $l_{\text{art}}$, $m_{\text{art}}$ and $m_{\text{base}}$ indicate the distance from wrist to the articulation axis and the weights of the object's parts, respectively. We ablate this component $\textbf{I}_{\text{art}}$ in \secref{ablation}. 

The goal features $\mathcal{G}$ guide the policy towards the hand pose reference and the target articulation angle. They are defined as: $\mathcal{G}=(\tilde{\textbf{g}}_q, \tilde{\textbf{g}}_x, \textbf{g}_c, \textbf{g}_a)$. In particular, $\tilde{\textbf{g}}_q = \overline{\textbf{q}}-\textbf{q}$ is the distance between the target and the current hand joint rotations (including wrist). The term $\tilde{\textbf{g}}_x = \overline{\textbf{x}}-\textbf{x}$ is the distance between the target and the current hand joint position, which can be computed from the hand pose using forward kinematics. $\textbf{g}_c=[\overline{\textbf{c}}|\overline{\textbf{c}}-\textbf{c}]$ contains the target contacts and the difference between the target and the current binary contact vector. $\textbf{g}_a=\overline{\omega}-\omega$ is the difference between the the target and the current object articulation angle. The target position, pose, and contacts are extracted from the hand pose reference. The target articulation angle is set to zero for grasping and otherwise set to a random angle during training. The goal features are expressed in either the object's base or articulation coordinate frame, depending on the part that needs to be manipulated.

\subsection{Reward Function}
The individual time-step reward function should guide our policy towards a solution that imitates the reference pose and fulfills the task objectives at the same time. Therefore, we define it as follows:
\begin{equation}
    r = r_{\text{im}} + r_{\text{task}}, 
\end{equation}

\noindent where $r_{\text{im}}$ is the reward for imitating the reference pose and $r_{\text{task}}$ contains the task objective. The imitation reward is defined as:
\begin{equation}
    r_{\text{im}} = r_p + r_c + r_{\text{reg}}.
\end{equation}
The pose reward $r_p$ considers both the joint position and joint angle error. The joint position error is the weighted sum of the distances between target and current positions $\overline{\textbf{x}}$ and $\textbf{x}$ of every joint, and the joint angle error measures the L2-norm of the differences between the target and current finger joint (and wrist) angles $\overline{\textbf{q}}$ and $\textbf{q}$:

\begin{equation}
    r_p = -\sum_{i=1}^Lw_{px}^i||\overline{\textbf{x}}^i-\textbf{x}^i||^2 - w_{pq}||\overline{\textbf{q}}-\textbf{q}||,
\end{equation}
\noindent where $w_{px}^i$ and $w_{pq}$ are weights for the respective terms.
The contact reward $r_c$ is composed of a relative contact term, which corresponds to the fraction of target contacts $\overline{\textbf{c}}$ the hand has achieved, and a contact impulse term, which encourages the amount of force $f$ applied on desired contact joints, capped by a factor proportional to the object's weight $m_o$:

\begin{equation}
    r_c = w_{cc}\frac{\overline{\textbf{c}}^T\textbf{I}_{f>0}}{\overline{\textbf{c}}^T\overline{\textbf{c}}} + w_{cf} min(\overline{\textbf{c}}^T\textbf{f}, \lambda m_o),
\end{equation}

where $w_{cc}$ and $w_{cf}$ indicate the respective weights. The term $r_{\text{reg}}$ regularizes the linear and angular velocities of the hand and object:
\begin{equation}
    r_{\text{reg}} = -w_{rh}||\dot{\textbf{T}}||^2 - w_{ro}||\dot{\mathbf{\Omega}}||^2.
\end{equation}

\noindent The task reward $r_{\text{task}}$ consists of two incentives: opening the object to a target articulation angle and avoiding the movement of the object base from its initial pose:

\begin{equation}
    r_{\text{task}} = -w_{tq}||\overline{\omega}-\omega|| -w_{tx}||\textbf{p}^0-\textbf{p}||^2,
\end{equation}
where $\overline{\omega}$ and $\omega$ are the the target and the current articulation angle, $\textbf{p}^0$ and $\textbf{p}$ are the object's initial and current position. The weights $w_{tq}$ and $w_{tx}$ are used to balance the terms. All the weight values are reported in \suppl.

\subsection{Curriculum}
\label{curriculum}
Training our policies with non-stationary objects from the beginning makes it difficult to learn the precise control necessary for fine-grained articulation. To address this, we introduce a learning curriculum that consists of two phases. In the first phase, we fix the objects to the table surface and train each hand separately in its own physics environment (grey shaded box in \figref{fig:structure}). This lets the policies learn precise finger movements and articulation. It also enables faster training, since the physics simulation speed scales roughly quadratically with the number of contacts in the environment. In the second phase, we move to the more complex setting where the object base is not fixed to the surface and the hands are both simulated in the same environment (purple shaded box in \figref{fig:structure}). In this setting, the policies need to learn to articulate the object without moving the object base or even tipping the whole object over. Additionally, the hands must collaborate, i.e., one hand should grasp the object without moving it too much, such that the other hand can successfully manipulate the object. In \secref{ablation}, we ablate the effectiveness of our curriculum. 

\section{Sequence Generation}
\label{sequence}
Given the unified policy per hand that can grasp and articulate objects, we now solve the \task~task (see \secref{sec:task}) by combining the different subtasks. To achieve this, we use two pairs of hand pose references $\textbf{D}^{\text{grasp}}$ and $\textbf{D}^{\text{art}}$. In the first phase, the hand policies are executed until a stable grasp is reached. In this case, the target object articulation angle $\overline{\omega}$ is set to zero and $\textbf{D}^{\text{grasp}}$ is used as input. To move the object to its target 6D global pose, we use the policies to keep a stable grasp on the object and employ the motion synthesis module according to D-Grasp~\cite{christen2022dgrasp}. Note that in the case where the hand pose reference contains only single hand manipulation, we simply fix the other hand. After having relocated the object, we need to transition from grasping into pre-grasp poses for articulation. This is achieved through a heuristics-based control scheme. First, we release the grasps by bringing the fingers into open hand poses and moving them away from the object following the direction that points from the object center to the wrist. Next, we linearly interpolate a trajectory between the hand poses and pre-grasp poses for articulation $\textbf{D}^{\text{art}}$. The pre-grasp poses correspond to $\textbf{D}^{\text{art}}$ with a linear translation in global space. They are computed by setting them at a small distance away from direction of the object center to the wrist poses of the reference. In the last phase, we use our articulation policy to approach the object and articulate it to reach the target articulation angle.

\section{Experiments}
\label{sec:experiment}
We conduct several experiments to evaluate our framework. We first report experimental details in Section \ref{experiment}. We then conduct quantitative evaluations on grasping and articulation tasks in \secref{atomic} and \secref{sec:recon}, including experiments with imperfect hand pose references from images. Finally, we provide ablations to show the importance of our method's components in Section \ref{ablation}. Please see our \suppl video for qualitative examples.

\subsection{Experimental Details}
\label{experiment}

\noindent \textit{Implementation Details}
We use PPO~\cite{schulman2017proximal} for RL training and RaiSim~\cite{Hwangbo2018Raisim} for the physics simulation. We train all policies using a single Nvidia RTX 6000 GPU and 128 CPU cores. Training our method takes roughly three days. We will release all code and data for future research.\\

\noindent \textit{Dataset}
\label{dataset}
We utilize the ARCTIC dataset~\cite{fan2023arctic}, which contains fully annotated two-hand interaction sequences including dexterous grasping and manipulation of articulated objects. We separate all available sequences into the different interactions of grasping and articulation. For each interaction, we extract a single pair of hand pose references using heuristics (see \suppl for details). Since the ground-truth annotation for the test split is not released, we create a custom 65\%/35\% train/test-split over all sequences of the training and validation sets, with a total of 488 and 257 hand pose references for training and testing, respectively. For the experiments with image-based estimates, we only use the validation set consisting of 60 hand pose references, because the pose estimation model was trained on the ARCTIC training set (see \secref{sec:recon}).\\

\noindent \textit{Metrics} 
\label{metrics} 
We mostly follow related work~\cite{jiang2021graspTTA, christen2022dgrasp} and define three metrics for grasping (success rate, position and angle error), three metrics for articulation (success rate, simulated distance and angle error), and one additional metric for the \task~task. We omit the interpenetration metric since all of our baselines include a physics simulation which exhibits no interpenetration.

\noindent\textbf{Grasp Success Rate (Suc. G)}: A grasp is defined as success if the object is lifted higher than 0.1m and does not fall until the sequence terminates.

\noindent\textbf{Position Error (PE)}: The mean position error between the object's final and target 3D position in meters.

\noindent\textbf{Angle Error (AE)}: The mean angle error between the object's final and target base orientation measured as geodesic distance in radian.

\noindent\textbf{Articulation Success Rate (Suc. A)}: An articulation is defined as success if the hand can articulate the object for more than 0.3 rad and the articulated part does not slip until sequence termination.

\noindent\textbf{Articulation Angle Error (AAE)}: The mean error between the object final and target articulation angle in radian.

\noindent\textbf{Simulated Distance (SD)}: As articulation should not move the object base, we report average displacement of the object base in meters.

\noindent\textbf{Task Success Rate (Suc. T)}: We deem a task as success if the PE $< 0.05$m, the AE $<0.2$rad, and the AAE $<0.5$rad.\\

\begin{figure*}[t]
	\centering
 \includegraphics[width=0.93\textwidth]{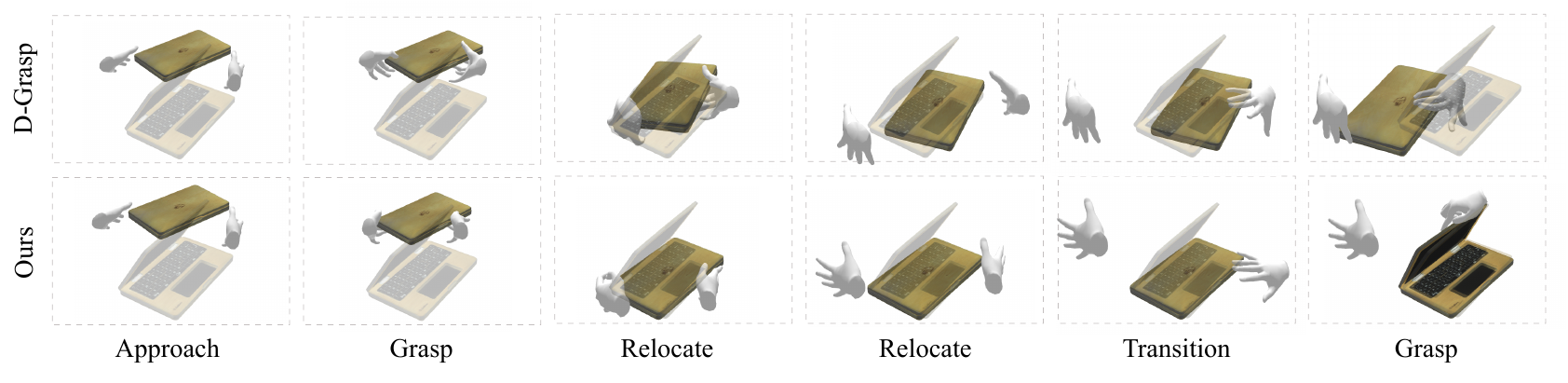}
\caption{\textbf{Qualitative evaluation of \task.} \dgrasp can grasp and relocate the object successfully, but fails to articulate the object. Ours is more successful at tackling this task and can articulate the object after relocation.}
\label{fig:comparison}
\end{figure*}

\noindent \textit{Baselines} 
\label{baseline}
D-Grasp is the most related to ours (see \tabref{tab:related_work})~\cite{christen2022dgrasp}, so we propose baselines following it:

\noindent\textbf{D-Grasp}: For grasping, we use vanilla D-Grasp and train the policies of the two hands directly with non-stationary objects. To compare D-Grasp to our method for articulation, we adjust the wrist control in D-Grasp. We first gradually increase the angle of the articulated joint and calculate the target 6D wrist pose with inverse kinematics by assuming that the wrist is fixed to the articulated part of the object. We then feed the wrist target pose to the PD controller.

\noindent\textbf{PD+IK}: We use the hand reference poses and set them as targets to the PD controller. The wrist for the articulation is controlled in the same way as in D-Grasp.

\begin{table}[t]
\centering
\vspace{-0.35cm}
\begin{minipage}{\columnwidth}
\begin{center}
\resizebox{1.0\linewidth}{!}{
\begin{tabular}{l|ccc|ccc}
   \toprule
    & \multicolumn{3}{c|}{Grasping} & \multicolumn{3}{c}{Articulation} \\
    \midrule
    Model & Suc. G $\uparrow$ & PE $\downarrow$ & AE $\downarrow$ & Suc. A $\uparrow$ & AAE $\downarrow$ & SD $\downarrow$ \\
    \midrule
    PD+IK                                             & 0.13 & 1.20 & 1.50 & 0.28 & 0.80 & 0.39 \\
             \dgrasp & \textbf{0.72} & \textbf{0.12} & \textbf{0.62} & 0.22 & 0.93 & 0.49 \\
             Ours & 0.71 & 0.13 & 0.69 & \textbf{0.55} & \textbf{0.57} & \textbf{0.01} \\
   \bottomrule
\end{tabular}
} 
\end{center}
\vspace{-0.4cm}
\end{minipage}
\caption{\heading{Quantitative comparison for grasping and articulation tasks} When the tasks are decoupled, we find that our method outperforms the baselines on articulation and performs comparably to D-Grasp on grasping.}
\label{tab:one}
\vspace{-0.35cm}
\end{table}

\subsection{Evaluation}
\label{atomic}

We first evaluate grasping and articulation tasks separately and then conduct experiments on the \task~task (see \secref{sec:task}).\\

\begin{figure}[t]
	\centering
 \includegraphics[width=0.47\textwidth]{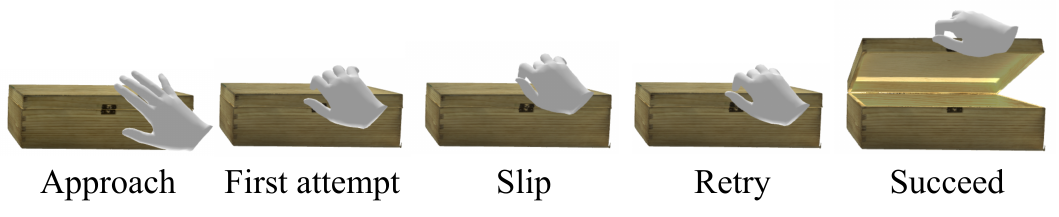}
  \vspace{-0.2cm}
\caption{\textbf{Qualitative articulation result.} The hand shows some recovery ability from failure cases. Zoom in for details.}
\label{fig:recovery}
\end{figure}

\noindent \textit{Grasping and Articulation} For grasping, we pre-sample 30 \sixD target object poses randomly (see \suppl for details). To control the wrist movement for relocation after grasping the object with our method and the PD+IK baseline, we adopt the same motion synthesis module as in~\cite{christen2022dgrasp} (see \secref{sequence}). 
For articulation, we evaluate each hand pose reference on 5 target articulation angles: $\{0.5, 0.75, 1.0, 1.25, 1.5\}\ \text{rad}$. For both tasks, the initial hand poses are set at a pre-defined distance away in the direction that points from the object center to the wrist of the hand pose references, with partially opened hands. The quantitative results are shown in \tabref{tab:one}. Our method significantly outperforms the PD+IK baseline on both grasping and articulation. Our policy has considerably better articulation performance and comparable grasping performance compared with D-Grasp. The results also show the difficulty of articulation and indicate that our learning-based wrist control is favorable for articulation compared to D-Grasp's non learning-based approach. Furthermore, as shown in the qualitative result \figref{fig:recovery}, we observe some recovering behavior from failure cases, which indicates the robustness of our policy. In particular, the agent fails to grasp first but tries again to find a better grasp until it succeeds in articulating the object. Qualitative comparisons and more examples of generated sequences are presented in \suppl figures. To evaluate the generalization ability of our framework, we conduct a proof-of-concept experiment with a single left-out object. The result indicates that our framework can generalize to an unseen object with about 15\% performance drop. We hypothesize that this is because the hand pose reference serves as a strong prior to the policy. However, more thorough evaluations need to be carried out once accurate 3D datasets with more articulated objects and hand-object poses become available. \\

\begin{table}[t]%
\vspace{-0.35cm}
\begin{minipage}{\columnwidth}
\begin{center}

\resizebox{0.7\linewidth}{!}{
\begin{tabular}{l|cccc}
   \toprule
    Models & Suc. T $\uparrow$ & PE $\downarrow$ & AE $\downarrow$ & AAE $\downarrow$  \\
   \midrule
    \dgrasp     & 0.11 & 0.05 & 0.15 & 0.66   \\
    Ours        & \textbf{0.50} & \textbf{0.03} & \textbf{0.10} & \textbf{0.41}   \\
   \bottomrule
\end{tabular}
}
\end{center}
\vspace{-0.4cm}
\end{minipage}
\caption{\heading{Evaluation for our \textit{Dynamic Object Grasping and Articulation} task} Our method outperforms D-Grasp on all metrics when evaluated on the task of transitioning an articulated object into a target articulated object pose.}
\label{tab:main_task}
\vspace{-0.35cm}
\end{table}%

\noindent \textit{\task} To evaluate this task (see \secref{sec:task}), we combine all grasping hand pose references with all articulation hand pose references per object, and sample a random target articulated object pose per trial (see \suppl for details). This results in roughly 7500 evaluation trials in total. We report the average pose errors and the task success rate in \tabref{tab:main_task}. 
Our method outperforms \dgrasp significantly in all metrics. For example, our method achieves a success rate of $5\times$  than that of D-Grasp. This shows that while D-Grasp can perform well in grasping when decoupled, it struggles in this composed task. We provide a qualitative comparison in \figref{fig:comparison} and a demonstration of a longer sequence with multiple objects in our \suppl figures. Moreover, see our \suppl video for more qualitative examples.

\subsection{Generation with Reconstructed Hand Pose}
\label{sec:recon}
\begin{figure}
\vspace{-0.15cm}
  \centering
    \includegraphics[width=0.99\linewidth]{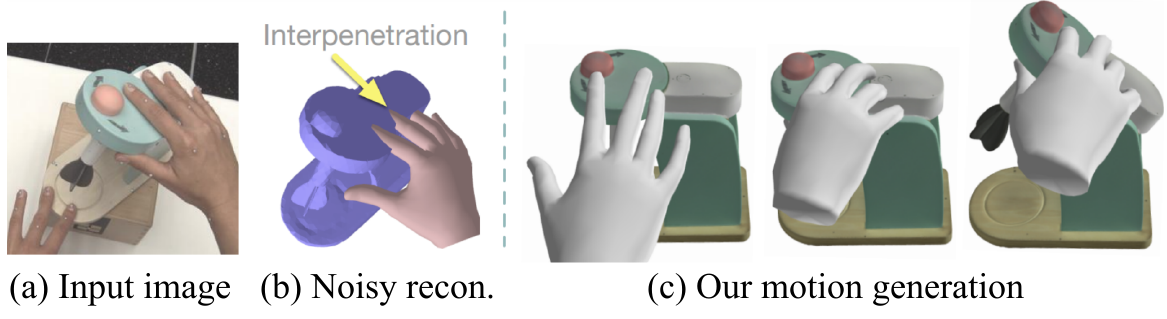}
  \vspace{-0.15cm}
  \caption{\textbf{Motion generation.} Our method can synthesize new motion sequences (c) with a noisy hand pose reference (b) reconstructed from a single RGB image (a).}
  \label{fig:reco}
  \vspace{-0.2cm}
\end{figure}

We now evaluate our method with hand pose references obtained from image predictions via the off-the-shelf hand-object pose regressor from \arctic. In particular, we estimate hand and object poses from images of the unseen validation subject in the ARCTIC and use the reconstructed results as input to our method and baselines. We separate the evaluation into grasping and articulation and present the results in \tabref{tab:recon}. Despite reconstruction noise such as hand-object interpenetration, our method can retain comparable performance as in the experiment with hand pose references from motion capture. This indicates our robustness to prediction noise and its potential to synthesize new motions with hand-object pose references from single images.
An example of our generated motion is shown in \figref{fig:reco}.

\subsection{Ablations}
\label{ablation}
We ablate the impact of the newly introduced components on our framework. To this end, we compare our full method against i) training both hands cooperatively and with a non-stationary object from the start of training (\textit{w/o curriculum}) ii) training the hands separately and with a fixed-base object (\textit{w/o cooperation}). Additionally, we train our method without the articulation features $\textbf{I}_{art}$ (\textit{w/o art. features}, cf. \secref{method}). The results are presented in \tabref{tab:two}. Without the curriculum, the policy achieves slightly better performance for grasping, but struggles with articulation. This is because grasping has different wrist motion with articulation which is easier to be learnt, which indicates the importance of a controlled setting to learn fine-grained articulation first. When training the hands separately without cooperation, grasping performance decreases because the hands cannot learn to collaborate for two-handed grasping. Lastly, the articulation features $\textbf{I}_{art}$ improve all articulation metrics, indicating that it provides important information about the object to the policy. 
\begin{table}%
\begin{minipage}{\columnwidth}
\begin{center}
\resizebox{1.0\linewidth}{!}{
\begin{tabular}{l|ccc|ccc}
   \toprule
   & \multicolumn{3}{c|}{Grasping} & \multicolumn{3}{c}{Articulation} \\
    \midrule
    Models & Suc.G $\uparrow$ & PE $\downarrow$ & AE $\downarrow$ & Suc.A $\uparrow$ & AAE $\downarrow$ & SD $\downarrow$\\
   \midrule
    \dgrasp   & 0.60 & \textbf{0.16} & \textbf{0.78} & 0.20 & 1.07 & 0.63  \\
    Ours    & \textbf{0.64} & \textbf{0.16} & 0.80 & \textbf{0.54} & \textbf{0.55} & \textbf{0.01}   \\
    \midrule
    Ours$^{*}$        & 0.67 & 0.14 & 0.95 & 0.54 & 0.53 & 0.01  \\ 
   \bottomrule
\end{tabular}
}  

\end{center}
\vspace{-0.4cm}

\end{minipage}
\caption{\heading{Results with reconstructed hand pose references} When evaluated with predictions from images, we observe a minor drop in performance for grasping and articulation compared to mocap data. However, the overall performance shows that our method can handle noisy estimates. The asterisk (*) denotes using hand pose references from mocap. }
\label{tab:recon}
\end{table}%

\begin{table}%
\begin{minipage}{\columnwidth}
\begin{center}

\resizebox{1.0\linewidth}{!}{
\begin{tabular}{l|ccc|ccc}
   \toprule
    & \multicolumn{3}{c|}{Grasping} & \multicolumn{3}{c}{Articulation} \\
    \midrule
    Models & Suc. G $\uparrow$ & PE $\downarrow$ & AE $\downarrow$ & Suc. A $\uparrow$ & AAE $\downarrow$ & SD $\downarrow$ \\
   \midrule
    w/o curriculum     & \textbf{0.74} & \textbf{0.13} & \textbf{0.65} & 0.36 & 0.77 & 0.02  \\
    w/o cooperation    & 0.21 & 0.32 & 1.43 & 0.48 & 0.65 & 0.02 \\
    w/o art. features  & 0.67 & 0.15 & 0.73 & 0.48 & 0.67 & 0.01 \\
    \midrule
    Ours             & 0.71 & \textbf{0.13} & 0.69 & \textbf{0.55} & \textbf{0.57} & \textbf{0.01} \\
   \bottomrule
\end{tabular}
}

\end{center}
\vspace{-0.4cm}
\end{minipage}
\caption{\heading{Ablations} We ablate our curriculum, cooperative training, and the articulation features. All components are important aspects to achieve grasping and articulation with a single policy.}
\label{tab:two}
\vspace{-0.35cm}
\end{table}%

\section{Discussion and Conclusion}
We present a method to synthesize physically plausible bi-manual grasping and articulation of objects with a single policy. We introduce an RL-based method that learns hand-object interactions in a physics simulation from static hand pose references. To address the difficulty in learning precise control for articulation, we extract articulation features and propose a curriculum with increasing task difficulty. We show our method presents a first step towards the \task~task. Furthermore, we demonstrate that noisy hand-object pose estimates obtained from individual RGB images can be used as input to our method. In a proof-of-concept with a single left-out object we have shown that our policy has the potential to generalize to unseen objects, and better generalization may be achieved in the future when larger and more diverse datasets become available. A limitation of our method is that it sometimes generates unnatural poses caused by noisy hand pose references and the trade-off between the task and imitation reward, which is shown in our \suppl figures. This may be improved by integrating bio-mechanical constraints or hand pose priors obtained from data-driven methods into our framework. And generation without reference poses would be an interesting direction for the future work.
{
    \small
    \bibliographystyle{ieeenat_fullname}
    \bibliography{main}
}
\appendix
\clearpage
\setcounter{page}{1}
\maketitlesupplementary

The supplementary material contains this document and a video. We describe implementation details in \secref{supp:impl} and experimental details in \secref{supp:exp_details}. In \secref{supp:experiments}, we provide additional experiments. We will release all code and pre-trained models.

\section{Implementation Details}
\label{supp:impl}
For training, we use PPO \cite{schulman2017proximal} and follow the implementation provided in~\cite{christen2022dgrasp}. We present an overview of the important parameters and weight values of the reward function in \tabref{tab:params}. 

\begin{table}[ht]
\centering
\resizebox{0.8\columnwidth}{!}{
\begin{tabular}{ll}
\textbf{Hyperparameters PPO} & \textbf{Value} \vspace{0.1cm}\\
\hline\\
Epochs & 1e4\\
Steps per epoch &6e5\\
Environment steps per episode & 300 \\
Batch size & 2000 \\
Updates per epoch & 20 \\
Simulation timestep & 2.5e-3s \\
Simulation steps per action & 4 \\
Discount factor $\gamma$ & 0.996 \\
GAE parameter $\lambda$ & 0.95 \\
Clipping parameter &0.2 \\
Max. gradient norm & 0.5 \\
Value loss coefficient & 0.5\\
Entropy coefficient & 0.0\\
Optimizer & Adam~\cite{adam2015}\\
Learning rate & 5e-4\\
Hidden units & 128 \\
Hidden layers & 2 \\
\\
\textbf{Weight Parameters} & \textbf{Value} \vspace{0.1cm}\\
\hline\\
$w_{px}$ & 3.0\\
$w_{px,\text{fingertip}}$ & 12.0\\
$w_{pq}$ & 0.2\\
$w_{cc}$ & 1.5\\
$w_{cf}$ & 1.5\\
$w_{rh}$ & 0.5\\
$w_{ro}$ & 0.2\\
$w_{tq}$ & 1.5\\
$w_{tx}$ & 0.2\\
$\lambda$ & 5.0\\
\end{tabular}
}
\captionof{table}{Hyperparameters of our RL algorithm and the weight values of the reward function. }
\label{tab:params}
\end{table}

\section{Experimental Details}
\label{supp:exp_details}
\subsection{Dataset}
In our experiments, we use the ARCTIC dataset~\cite{fan2023arctic} and include sequences from all training subjects and the recently released data from the validation subject s05. We exclude the three objects "scissors", "capsule machine", and "phone" from our experiments. "Scissors" is different from all other objects as it cannot be split into a clear base and articulation part and requires in-hand manipulation. "Phone" and "capsule machine" have very small and thin articulation parts which cannot be modeled with our method currently. We then extract hand pose references according to \secref{supp:label_gen}. From these references, we create a 65\%/35\% train/test-split. 
In total, we generate 745 hand pose references for 8 objects from the dataset, with a train/test split of 488/257 hand pose references.

\subsection{Hand Pose Reference Generation}
\label{supp:label_gen}
We now describe the procedure of retrieving hand pose references from motion capture sequences. Since the sequences contain several different interactions of object manipulation, from which a lot of hand pose references could be extracted, we devise heuristics to obtain diverse frames and avoid redundancy. We distinguish between two types of manipulation in this paper: grasping and articulation.

For each sequence, we first remove all frames where none of the hands is in contact with an object. Next, we filter all remaining frames for grasping and articulation. An interaction is determined as grasping if an object is moved from its underlying surface, i.e., if the velocity of the object base $\dot{\textbf{B}}$ is higher than a threshold $\epsilon_v$. On the other hand, if the articulation angle $\omega$ is changed, we deem an interaction as articulation. To avoid redundancy in hand pose reference frames, we make the assumption that the hand pose does not drastically change during one interaction. Hence, we choose one frame per interaction subsequence.

\begin{figure}
  \centering
  \begin{subfigure}{1.0\linewidth}
    \centering
    {\includegraphics[width=0.3\textwidth]{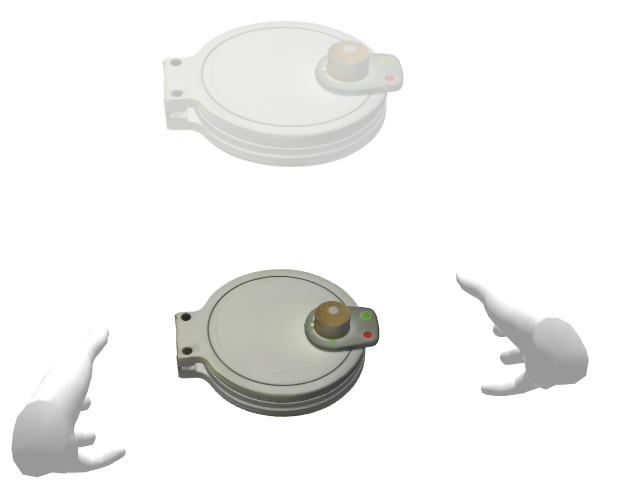}
     \includegraphics[width=0.3\textwidth]{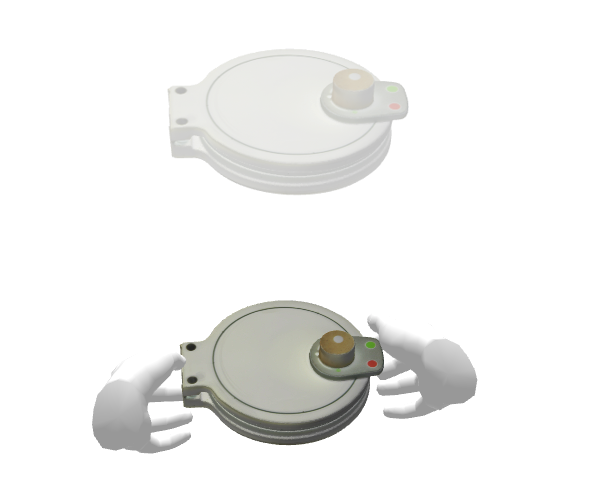}
     \includegraphics[width=0.3\textwidth]{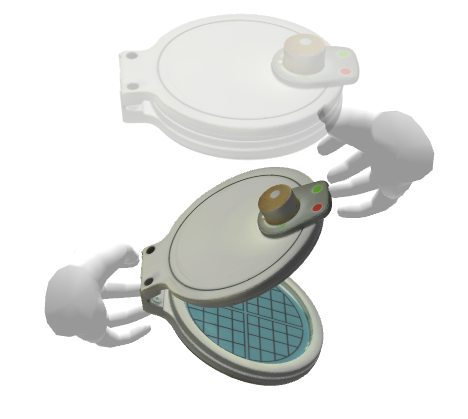}}
    \caption{PD+IK}
  \end{subfigure}
  \hfill
  \begin{subfigure}{1.0\linewidth}
        \centering
    {\includegraphics[width=0.3\textwidth]{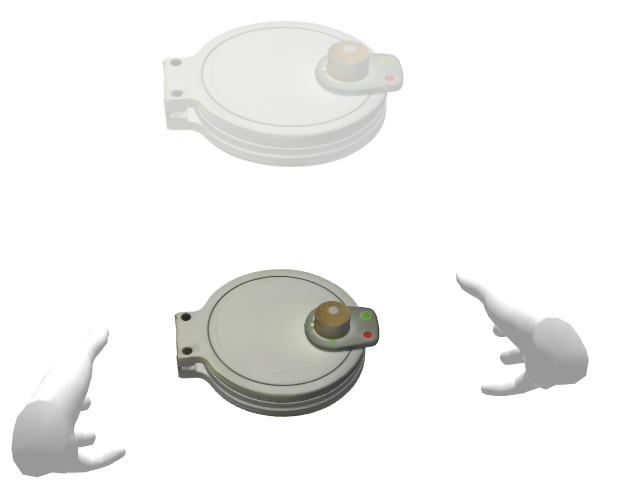} 
     \includegraphics[width=0.3\textwidth]{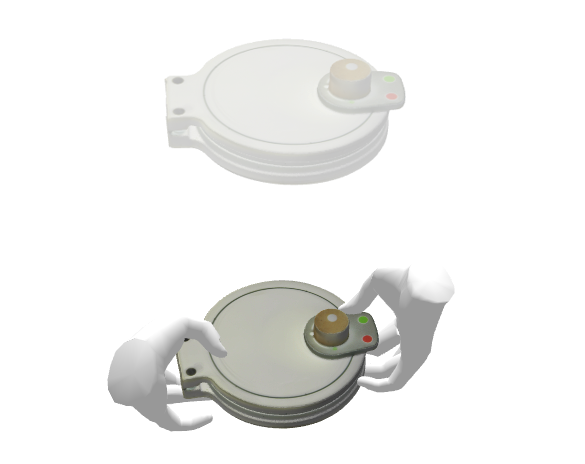}
     \includegraphics[width=0.3\textwidth]{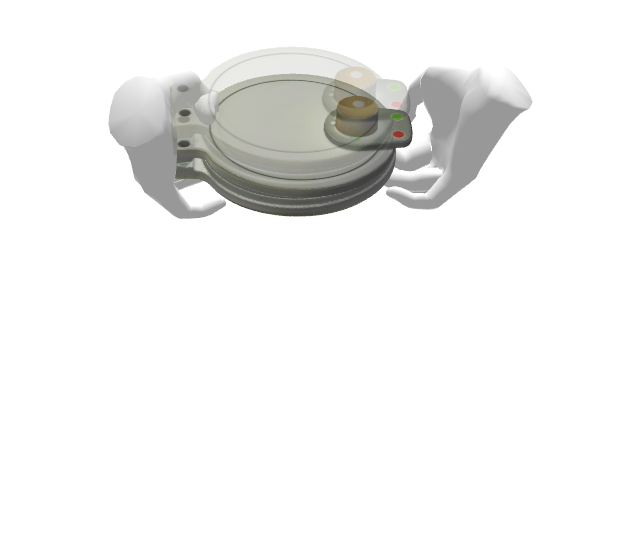}}
    \caption{D-Grasp}
  \end{subfigure}
  \hfill
  \begin{subfigure}{1.0\linewidth}
        \centering
        {\includegraphics[width=0.3\textwidth]{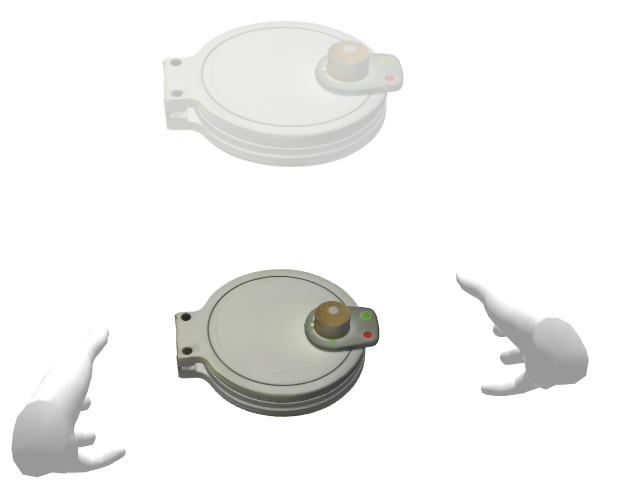}
     \includegraphics[width=0.3\textwidth]{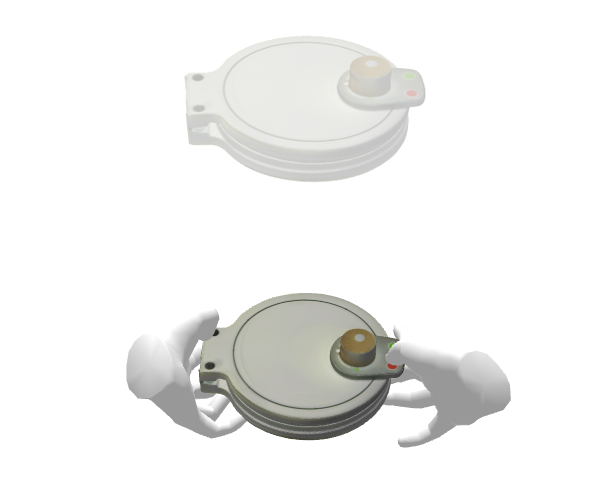}
     \includegraphics[width=0.3\textwidth]{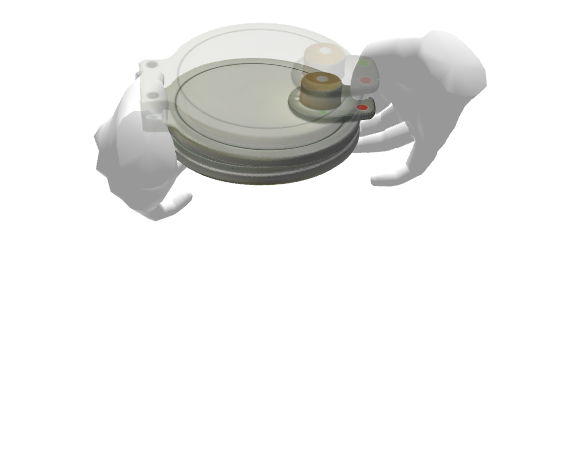}}
    \caption{Ours}
  \end{subfigure}
  \caption{\textbf{Qualitative evaluation of grasping.} When evaluated only on grasping, PD+IK often fails to successfully grasp the object. On the other hand, D-Grasp and ours succeed at the task.}
    \label{fig:comparison_grasp}
\end{figure}

\begin{figure}
  \centering
  \begin{subfigure}{1.0\linewidth}
        \centering
        {\includegraphics[width=0.3\textwidth]{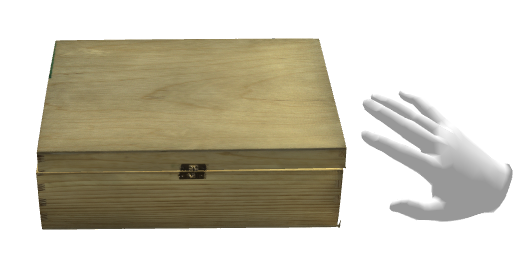} 
     \includegraphics[width=0.3\textwidth]{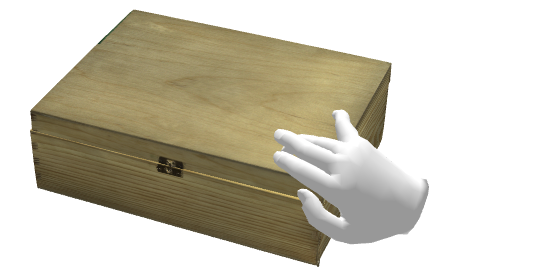}
     \includegraphics[width=0.3\textwidth]{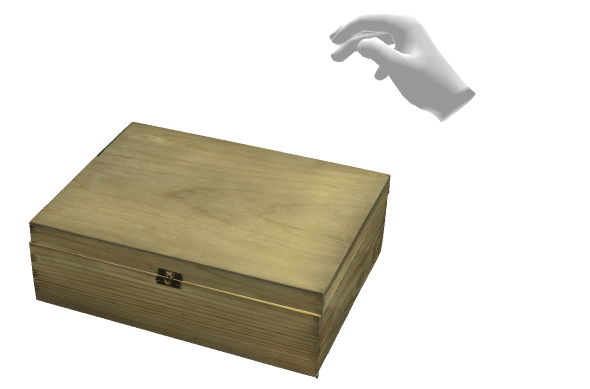}}
    \caption{PD+IK}
  \end{subfigure}
  \hfill
  \begin{subfigure}{1.0\linewidth}
        \centering
        {\includegraphics[width=0.3\textwidth]{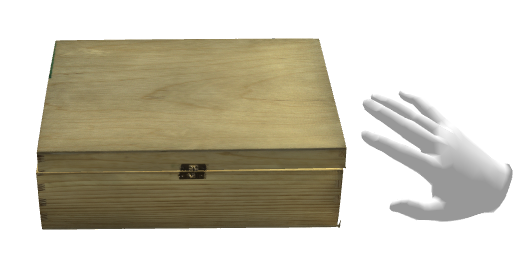} 
     \includegraphics[width=0.3\textwidth]{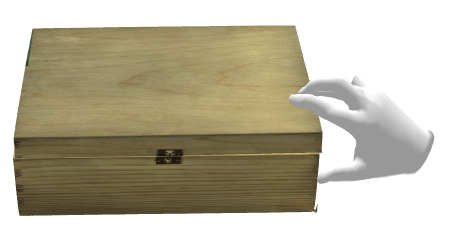}
     \includegraphics[width=0.3\textwidth]{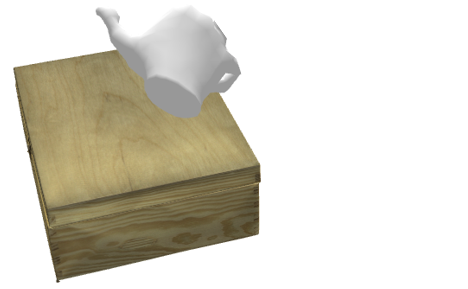}}
    \caption{D-Grasp}
  \end{subfigure}
  \hfill
  \begin{subfigure}{1.0\linewidth}
        \centering
        {\includegraphics[width=0.3\textwidth]{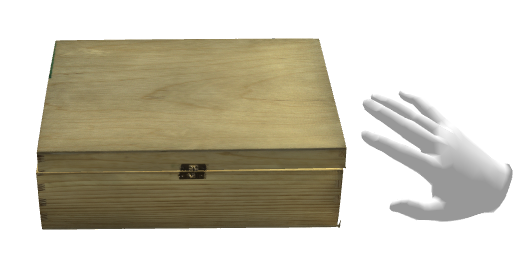} 
     \includegraphics[width=0.3\textwidth]{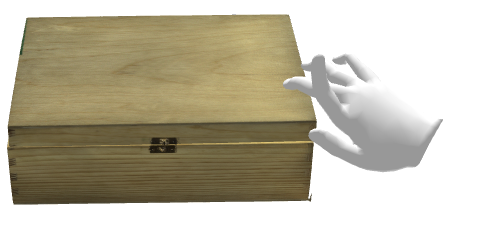}
     \includegraphics[width=0.3\textwidth]{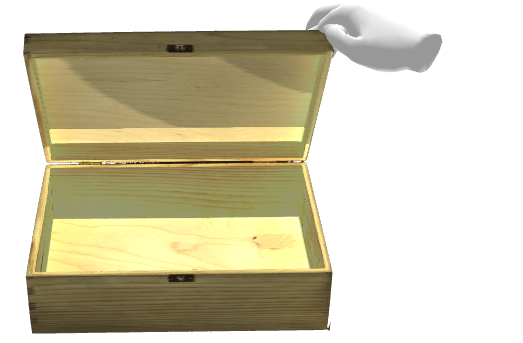}}
    \caption{Ours}
  \end{subfigure}
  \caption{\textbf{Qualitative evaluation of articulation.} When evaluated only on articulation, both PD+IK and D-Grasp often fail at the task. On the other hand, our method can articulate the object successfully.}
    \label{fig:comparison_arti}
\end{figure}

\subsection{Grasping and Articulation}
To evaluate grasping, we generate 30 target \sixD object poses for each hand-pose reference. The target positions are sampled within a range of [-0.15m, 0.15m] in x and y directions and [0.15m, 0.45m] for the z direction. The target object orientation is the initial object orientation disturbed with noise in the range of [-0.3rad, 0.3rad] for all rotation axes. To evaluate articulation, we set 5 target joint angles per trial: 0.5rad, 0.75rad, 1.0rad, 1.25rad and 1.5rad. 

\subsection{Dynamic Object Grasping and Articulation}
For the evaluation of \task, we randomly sample the target articulated object poses $\overline{\mathbf{\Omega}}$, which consists of the target 6D base pose and the target object joint articulation angle. The target base position is sampled within a range of [-0.1m, -0.05m] in x and y directions and 0m in z direction, since the objects should be relocated back onto the table. The target base orientation is the initial object orientation disturbed with noise in the range of [-0.4rad, 0.4rad] for the yaw axis. The target articulation angle is randomly sampled in the range of [0.5rad, 0.6rad].

\subsection{Hand Pose Reconstruction}
In this experiment, we use the pretrained image-based reconstruction model of ARCTIC~\cite{fan2023arctic} to predict hand-object poses from single images. Since their model is trained on the full training set and the test data is not released, we use the validation set (subject s05) to evaluate this experiment. This allows us to do a direct comparison of individual hand pose references from motion capture and image-based predictions. Our evaluation is conducted on 60 hand pose references selected with the heuristics explained in \secref{supp:label_gen}. We retrieve the images at the corresponding timesteps and pass them to the image-based prediction model. 

\begin{figure}
	\centering
\includegraphics[width=0.47\textwidth]{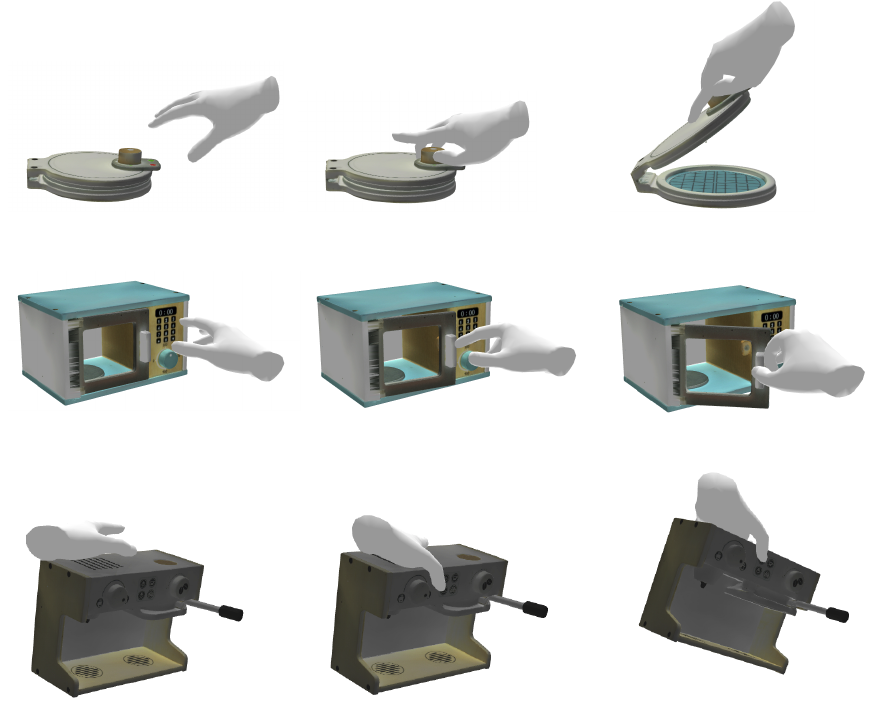}
\caption{\textbf{Qualitative outputs of our method.} We provide more sequences for grasping and articulation, which are generated by our method with a single pair of hand pose reference label per interaction. Each sequence is shown from left to right.}
\label{fig:more_samples}
\end{figure}

\section{Additional Experiments}
\label{supp:experiments}

We provide additional qualitative comparisons of our method with baselines for grasping and articulation in \figref{fig:comparison_grasp} and \figref{fig:comparison_arti}, respectively. Given a pair of policies (one per hand), our method can generate diverse grasping and articulation sequences across different objects, which is shown in \figref{fig:more_samples}. Please see our \suppl video for more qualitative examples.

\begin{figure*}[th]
	\centering
\includegraphics[width=1.0\textwidth]{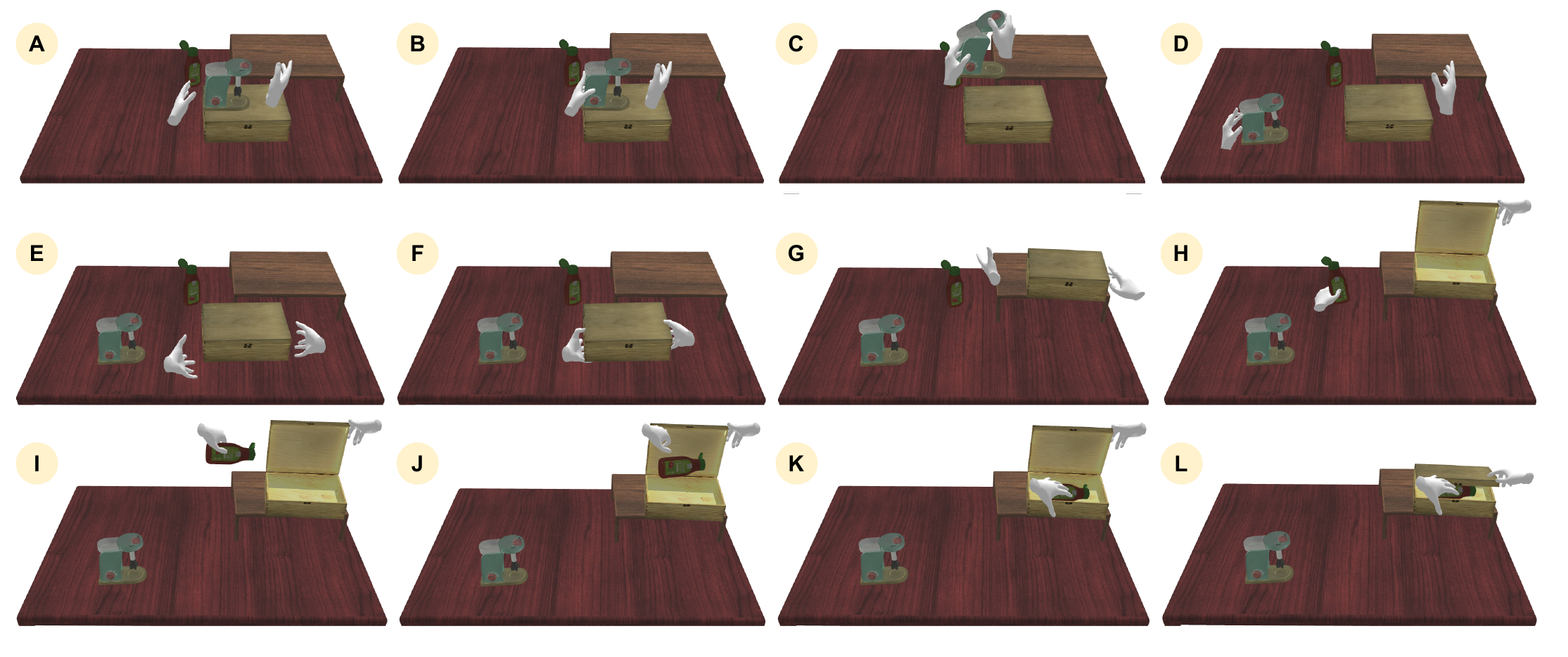}

\caption{\textbf{Long sequence with multiple objects.} We show that our method can generate sequences of manipulating multiple objects. (A) Approaching the mixer with the left hand. (B) Grasping the mixer with the left hand. (C) Articulating the mixer with the right hand while the left hand is holding it. (D) Putting the mixer down on the table. (E) Approaching the box with both hands. (F) Grasping the box with both hands. (G) Relocating the box on the table and moving the left hand to the ketchup bottle. (H) Grasping the ketchup bottle with the left hand and opening the box with the right hand. (I) Relocating the ketchup bottle while the box is being held open. (J) Dropping the ketchup bottle into the box. (K) Moving the left hand away from the box. (L) Closing the box with the right hand.}
\label{fig:long_sequence}
\end{figure*}

\begin{figure}
  \centering
  \begin{subfigure}{1.0\linewidth}
    \centering
     {\includegraphics[width=1.0\textwidth]{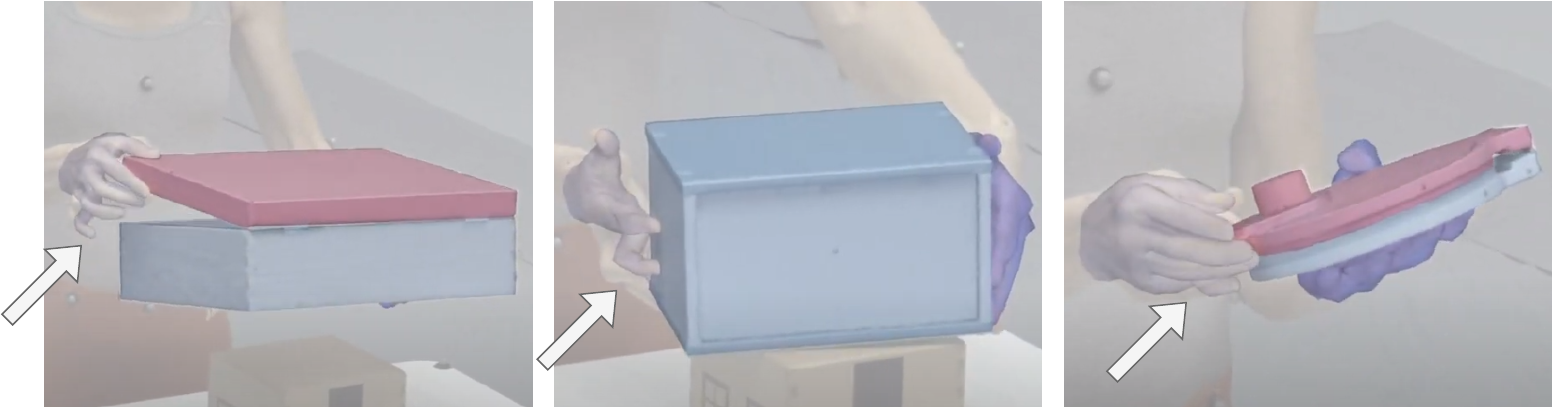}}
    \caption{Unnatural hand pose references}
    \label{fig:unnatural_a}
  \end{subfigure}
  \hfill
  \begin{subfigure}{1.0\linewidth}
    \centering
    {\includegraphics[width=1.0\textwidth]{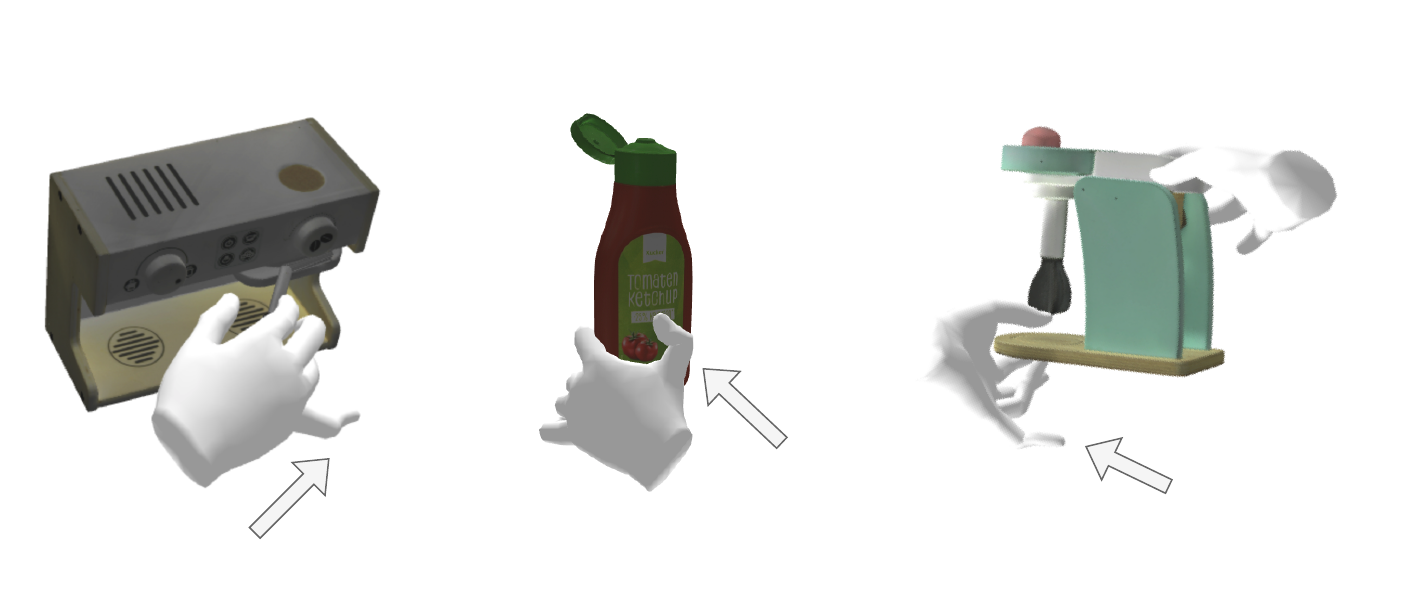}}
    \caption{Unnatural generated hand poses}
    \label{fig:unnatural_b}
  \end{subfigure}
  \caption{\textbf{Unnatural hand poses} (a) Some of the hand pose references we extract from the ARCTIC dataset contain unnatural hand poses. (b) Our method can output some unnatural hand poses, which can be due to noise in the hand pose references or because of the trade-off in the task objective.}
    
\end{figure}

\subsection{Long Sequence with Multiple Objects}
Our method can generate long motion sequences in environments with multiple objects, which is shown in \figref{fig:long_sequence}. We use a heuristics-based planner to compose the sequences. Learning a high-level planning module to couple the different phases is an interesting direction to explore in the future. Note that while we propose a controlled setting to evaluate the \task~task, the order of manipulations can also be reversed. For example, an object can first be articulated, and then be moved to a different location.

\subsection{Unnatural Poses}
As shown in \figref{fig:unnatural_b}, our method can generate unnatural poses, which we argue occurs because of noisy pose references from \arctic as seen in \figref{fig:unnatural_a}. We find that especially the index finger is often poorly labeled in the data, which translates to our policies. Developing hand pose priors to incentivize natural poses could be one way to mitigate this issue.

\end{document}